\documentclass[conference,a4paper]{IEEEtran}
\IEEEoverridecommandlockouts
\usepackage{cite}
\usepackage{amsmath,amssymb,amsfonts}
\usepackage{algorithmic}
\usepackage{graphicx}
\usepackage{textcomp}
\usepackage{tabularx,booktabs}
\usepackage[colorlinks=true]{hyperref}
\usepackage{eso-pic}

\usepackage{xcolor}
\def\BibTeX{{\rm B\kern-.05em{\sc i\kern-.025em b}\kern-.08em
    T\kern-.1667em\lower.7ex\hbox{E}\kern-.125emX}}
\graphicspath{ {./images/} }

\begin{document}
 \AddToShipoutPicture*{\small \sffamily\raisebox{1.2cm}{\hspace{1.8cm}Accepted, DICTA 2019}}

\title{LiteSeg: A Novel Lightweight ConvNet for Semantic Segmentation
}

%
%
%

\author{\IEEEauthorblockN{Taha Emara, 
Hossam E. Abd El Munim, Hazem M. Abbas}
\IEEEauthorblockA{Computer and Systems Engineering Department, Faculty of Engineering, Ain Shams University,\\ Cairo, Egypt \\ 
Email: \{taha@emaraic.com, 
hossameldin.hassan@eng.asu.edu.eg, 
hazem.abbas@eng.asu.edu.eg\}
}}
\maketitle

\maketitle

\begin{abstract}
Semantic image segmentation plays a pivotal role in many vision applications including autonomous driving and medical image analysis. Most of the former approaches move towards enhancing the performance in terms of accuracy with a little awareness of computational efficiency. In this paper, we introduce LiteSeg, a lightweight architecture for semantic image segmentation. In this work, we explore a new deeper version of Atrous Spatial Pyramid Pooling module (ASPP) and apply short and long residual connections, and depthwise separable convolution, resulting in a faster and efficient model. LiteSeg architecture is introduced and tested with multiple backbone networks as Darknet19, MobileNet, and ShuffleNet to provide multiple trade-offs between accuracy and computational cost. The proposed model LiteSeg, with MobileNetV2 as a backbone network, achieves an accuracy of 67.81\% mean intersection over union at 161 frames per second with $640 \times 360$ resolution on the Cityscapes dataset. 
\end{abstract}


\section{Introduction}
\label{sec:intro}
Semantic image segmentation is defined as the assigning of every pixel in a given image to a specific categorical label. Semantic segmentation~\cite{long2015fully, deeplabv3plus, zhao2017pspnet,deeplabv3}, and similar to image classification~\cite{xie2017aggregated,resnet,szegedy2015going} and object detection~\cite{yolo,lin2017focal}, has seen considerable progress due to the employment of deep learning architectures, especially convolutional neural networks (CNN). This progress has resulted in a much better quality of real-world applications, such as autonomous driving, medical diagnosis~\cite{ronneberger2015u}, and aerial image segmentation~\cite{marmanis2016semantic}. 

Despite the high accuracy achieved by recent proposed architectures~\cite{deeplabv3plus,zhao2017pspnet} for semantic segmentation, they are not computationally efficient especially for the applications that are needed to be run on edge devices, such as autonomous driving cars, robots, or augmented reality kits. Numerous attempts have been investigated in providing lightweight semantic segmentation architectures, such as ERFNet~\cite{erfnet}, ESPNet~\cite{espnet}, Enet~\cite{enet}, CCC~\cite{ccc2}, and DSNet~\cite{dsnet}. Some of these lightweight architectures attempts aimed at obtaining real-time performance with a considerable reduction in network parameters which significantly causes a loss in accuracy measures~\cite{espnet, enet, rtseg,ccc2}. Other methods paid more attention to both accuracy and real-time performance which leads to gain a better real-time performance when compared to complex networks and a better accuracy than the first group~\cite{erfnet, dsnet}. 
\\
\textbf{Semantic segmentation.} The Fully Convolutional Network (FCN)~\cite{long2015fully} is a pivotal approach which paves the way to employ deep learning methods into the semantic segmentation problem. In FCN, a classification model such as, GoogleNet~\cite{szegedy2015going} or VGG~\cite{simonyan2014very} was used as an encoder to extract features from tested images and then these feature maps were upsampled to pixelwise dense predictions by cascaded layers of unpooling and deconvolution operations. Accuracy of this architecture was improved by using skip architecture, in which semantic information from deep layer and spatial information from the earlier layers were combined to get better results. Despite the breakthrough of FCN architecture, it has suffered from low resolution prediction. Many variant of FCN are proposed to solve this problem. For example, the work in~\cite{eigen2015predicting} proposed a multi-scale network which employed a different three scale to generate a fine, high resolution predictions. Another solution was proposed by ~\cite{noh2015learning}, in which a more complex deconvolution network was used to produce high resolution predictions instead of the used one in~\cite{long2015fully} which used a single bilinear interpolation layer. A different approach~\cite{chen2018deeplab} employed dilated convolution to increase the receptive field without any increase the in number of parameters and computational cost, followed by bilinear interpolation layers to scale up the feature maps to the input image size. Then a conditional random field (CRF)~\cite{krahenbuhl2011efficient} was used as a post processing to refine the result image.
PSPNet~\cite{zhao2017pspnet}, Deeplabv3~\cite{deeplabv3}, Deeplabv3+~\cite{deeplabv3plus} capture information at multiple scales by either applying pooling operations with different kernel size and they called it pyramid pooling module (PPM) or employing dilated convolution with different rate and that was called Atrous Spatial Pyramid Pooling (ASPP).
\\
\begin{figure*}[t]
    \centering
    \includegraphics[height=6.8cm, width=\textwidth]{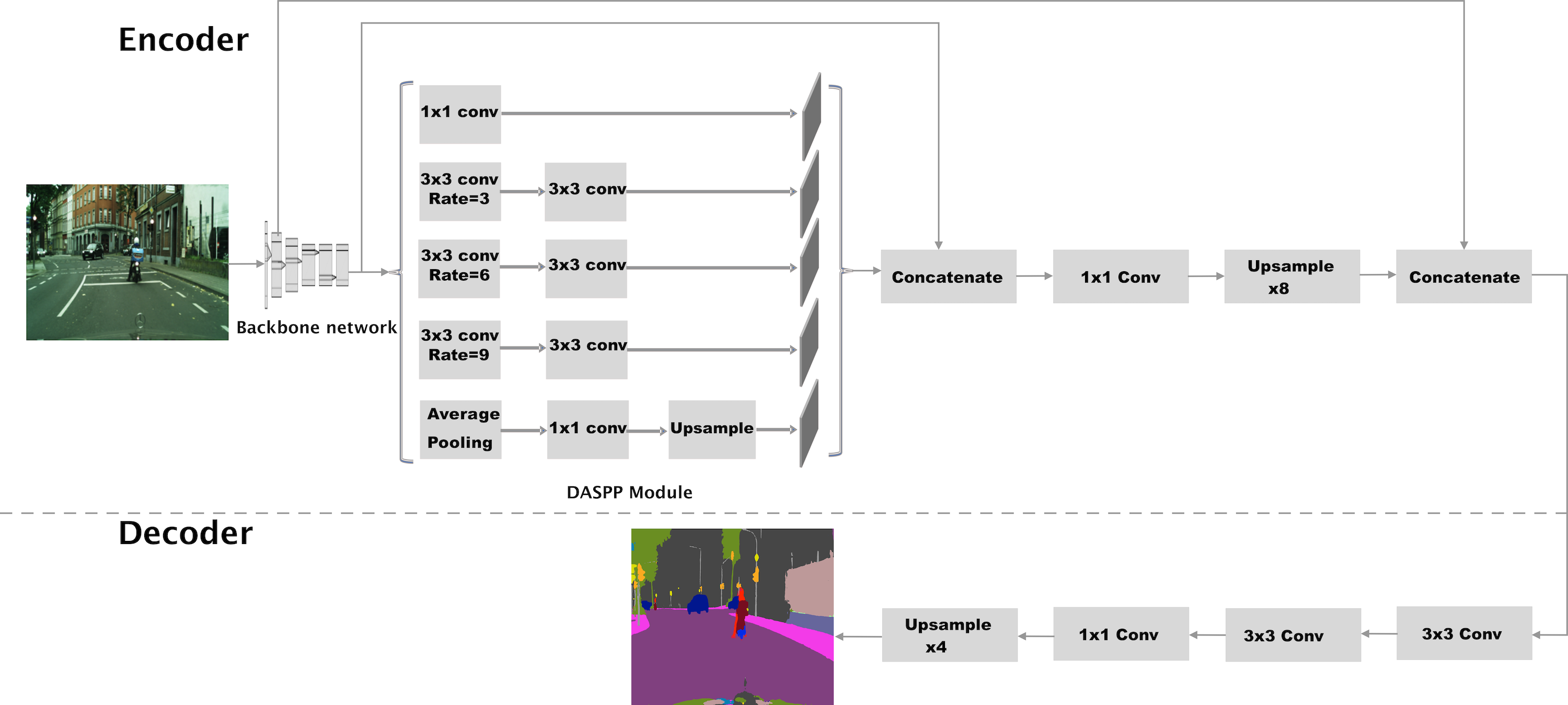}
    \caption{General LiteSeg diagram including the encoder module with its components backbone network and DASPP module, and the decoder module. Encoder module takes an input image and generates a high dimensional feature vector. The decoder module restores the spatial information from this feature vector.   }
    \label{fig:diagram}
\end{figure*}

\textbf{Real-Time Segmentation.} Most of the mentioned approaches are not efficient for real-time applications as they employed large backbone networks such as GoogleNet~\cite{szegedy2015going}, Xcepetion~\cite{simonyan2014very}, or ResNet~\cite{resnet}, or employed a large CNN architectures for both the encoder or decoder sides. This has lead to having a large number of parameters to be tuned and a large floating point operations (FLOPS), even though they are efficient form accuracy perspective.
Many approaches have been proposed to deal with this problem, e.g., ERFNet~\cite{erfnet} employed a residual connection and depthwise separable convolution to increase receptive field to achieve high accuracy with a reasonable performance. Alternatively, ESPNet~\cite{espnet} proposed an efficient module called efficient spatial pyramid (ESP), which uses point wise convolution and spatial pyramid of dilated convolution. ESPnet along with Enet provide a lightweight architectures but with a degradation in accuracy. RTSeg~\cite{rtseg} provided a decoupled encoder-decoder architecture which allows to plug any encoder (i.e., VGG16~\cite{simonyan2014very}, MobileNet~\cite{mobilenetv2}, ShuffleNet~\cite{shufflenet}, ResNet18~\cite{resnet}) or decoder (i.e., UNet~\cite{ronneberger2015u}, Dilation~\cite{yu2015multi}, SkipNet~\cite{long2015fully}) architectures independently. They have found out that using SkipNet architecture along with MobileNet and ShuffleNet provided the best trade-off between accuracy and performance.
\\Motivated by the encoder-decoder architecture, Atours Spatial Pyramid Pooling (ASPP), dilated convolution, and depthwise separable convolution, we design a novel architecture called LiteSeg which is capable of adapting any backbone network. This capability would allow a variety in trade-offs between computational cost and accuracy to fit multiple needs by choosing different backbone networks.
\\
In summary, our main contributions are:
\begin{itemize}
\item LiteSeg, a real time competitive architecture is presented and tested with three different backbone networks, Darknet19~\cite{yolo}, MobileNetV2~\cite{mobilenetv2}, and ShuffleNet~\cite{shufflenet}, achieving performance 70.75\%, 67.81\%, and 65.17\%, respectively on Cityscapes dataset.
\item A new deeper version of ASPP module is adapted to improve the results along with using long and short residual connection.
\end{itemize}
The rest of the paper is organized as follows. Section 2 describes the proposed architecture, LiteSeg, in details. In Section 3, both the accuracy and the computational efficiency of the proposed model is evaluated and the paper is finally concluded in Section 4.


\section{Methods}
\label{sec:1}
Here, we will describe our architecture LiteSeg and its new deeper version of ASPP module called Deeper atrous Spatial Pyramid Pooling (DASPP) module (Figure~\ref{fig:diagram}). In addition, the atrous convolution, depthwise separable convolution and long and short residual connection are briefly introduced. Then, Deeplabv3+~\cite{deeplabv3plus} which is used as the decoder module will be reviewed.
\subsection{Atrous Convolution}
\label{sec:2}
In convolutional architecture, decreasing the receptive field size will result in a spatial information loss that can be attributed to the strided convolution and pooling layers. To overcome this problem, the dilated convolution was used in~\cite{chen2018deeplab, yu2015multi} to increase the receptive field without any reduction in the feature map resolution and an increase in trainable parameters. This allows network to learn global context features across the entire image for refining full-resolution predictions. 

\subsection{Depthwise Separable Convolution}
\label{sec:3}
Standard convolution is computationally an expensive operation due to the large number of parameters to be tuned and thus the needed FLOPS. To tackle this problem, depthwise separable convolution is a suggested solution to replace standard convolution without compromising the accuracy.\\ 
The main idea of depthwise separable convolution is to split both the input and the kernel into channels -they share the same number of channels-, and each input channel will be convolved with the corresponding kernel channel. Then, the pointwise convolution is performed using a $1 \times 1$ kernel to project the output of the depthwise convolution into new channel space. Employing depthwise separable convolution~\cite{Nekrasov2018LightWeightRF} was empirically proven to reduce the computational cost with similar or better performance.
\begin{figure*}[h]
    \centering
    \includegraphics[width=0.90\textwidth ,height=8.6cm]{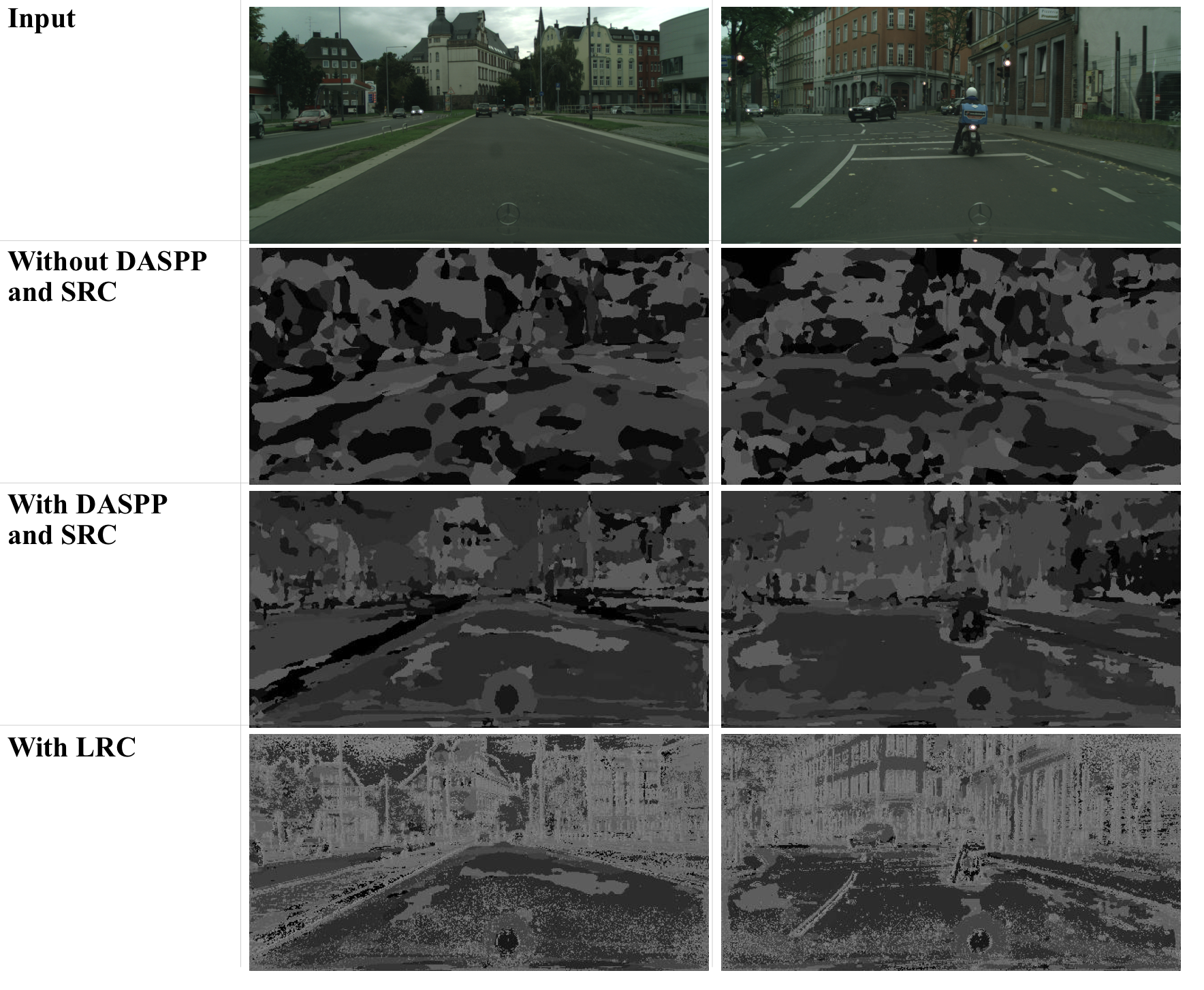}
    \caption{Visualization of the output after encoder module, to show the effectiveness of short residual connection (SRC), long residual connection (LRC), and DASPP module on our model performance. }
    \label{fig:mesh1}
\end{figure*}
\subsection{Long and short residual connection}
\label{sec:4}
He et al.~\cite{resnet} proposed a residual learning framework to allow training of very deep networks. Unlike the traditional feedforward neural network, ResNets introduce an identity shortcut connection. Let $X$ is the input feature map, $F(X)$ is the residual, and $H(X)$ is the output of residual block, the residual learning takes the form $\displaystyle H(X)=F(X)+X$ such that, if there is no residual it will work as identity mapping, that means it can eliminate the effect of the DASPP module if it turns out to be an unnecessary. The resulting learned residual assures that the proposed network would not perform worse than without it.\\
Fusion and reusing of low-level features -which include color blobs or edges- from bottom layers and high-level features from top layers have been proven to be helpful for high resolution segmentation~\cite{zhang2018exfuse}. This fusion can be done between feature maps from close layers by short residual connection (SRC) and far layers by long residual connection(LRC). These connections act as memory units~\cite{densenet} in the network as they allow preserving the information from the bottom layers to the top layers.\\
There are two approaches to carry out residual connections, one by element-wise addition \cite{resnet} and the other by concatenating the feature maps \cite{densenet}. Here, we employed the concatenation approach as an element-wise addition that require the residual output and the input have the same dimension width, height, and depth instead of the conventional concatenation which requires the same dimension of width and height only. The mismatch in width and height can be maintained by upsampling and optionally a $1 \times 1$ convolution can be used to reduce the depth of the features for computational efficiency. It was found out that long skip connection helps to make clearer semantic boundaries and short skip connections with DASPP help in fine tuning the semantic segments and thus providing richer geometrical information (Figure~\ref{fig:mesh1}).
\subsection{Proposed Encoder}
\label{sec:5}
The proposed encoder contains a backbone network architecture which acts as an image classification architecture for feature extraction. These architectures were chosen to meet our performance criteria, so we tested the architecture with different three lightweight models MobileNet, ShuffleNet, and Darknet19. Not only is the type of the backbone network controls the performance, but also the output stride~\cite{deeplabv3} which is defined as the ratio between the input image size and the last feature map of the encoder. Let height $H$, width $W$, and depth $C$ be the input image dimension and the outputs of the backbone network are $h, w,$ and $c$, so the output stride is defined as $\displaystyle os=H \times W/h \times w$. Decreasing the output stride leads to having high resolution feature map and also better results~\cite{deeplabv3} as  more spatial information throughout the network is preserved but it comes with computational cost. The output stride of backbone networks is controlled by removing max pool layers and modifying the stride for the last convolution layers. Deeplabv3+~\cite{deeplabv3plus} with output stride equal to 16 is the best trade-off between accuracy and computational efficiency. Moreover, they found out that the accuracy can be greatly improved using output stride equal to 8 but with huge computational cost and the computational efficiency can be improved by increasing the output stride to 32 with a compromise in accuracy. Therefore, the proposed backbone network is configured with output stride of 32 for MobileNetV2~\cite{mobilenetv2} and ShuffleNet~\cite{shufflenet} and output stride of 16 for Darknet19 to achieve different trade-offs between accuracy and speed. \\
DeepLabv3~\cite{deeplabv3} employs Atrous Spatial Pyramid Pooling (ASPP) module with different dilation rates to capture multi-scale information, following the presented approach in ParseNet~\cite{parsenet}. Here, a new deeper version of ASPP module is proposed (called Deeper Atrous Spatial Pyramid Pooling (DASPP)), by adding standard $3 \times 3$ convolution after $3 \times 3$ atrous convolutions to refine the features and also fusing the input and the output of the DASPP module via short residual connection. Also, the number of convolution filters of ASPP is reduced from 255 to 96 to gain computational performance.

\begin{table*}[t]
\caption{Evaluation results in mIOU on the Cityscapes validation set using {\it LiteSeg} with an input image size $512 \times 1024$ using different backbone networks. Baseline network is minimal version of Deeplabv3+. {\bf FT}: Using Coarse dataset. {\bf MS}: Multi-scale training strategy. {\bf DW}: Employing depthwise separable convolution. Results with '*' were evaluated on images with sizes $512\times1024$ and $1024\times2048$ and listed as $512\times 1024$ accuracy/1024x2048 accuracy.}
	\begin{center}
	\scalebox{1.3}{
		\begin{tabular}{@{}llllllll@{}}
	\toprule
Baseline & DASPP & FT & MS & DW & LiteSeg-Darknet & LiteSeg-MobileNet & LiteSeg-ShuffleNet \\ \midrule
   \checkmark   &      &    &    &    & 65.84\%         & 64.70\%           & 60.41\%            \\
         & \checkmark    &    &    &    & 68.21\%         & 64.80\%           & 61.3\%             \\
         & \checkmark   & \checkmark&    &    & 68.94\%/71.5\%*  & 66.4\%/67.8\%*     & 62.65\%/62.2\%*    \\
         & \checkmark    & \checkmark  & \checkmark  &    & 69.14\%/72.3\%*  & 66.49\%/69.3\%*    & 63.2\%/65.4\%*      \\
         &\checkmark   &\checkmark  & \checkmark & \checkmark  & 69.43\%/72.8\%*  & 66.48\%/70.0\%*    & 62.45\%/66.1\%*     \\ \bottomrule
		\end{tabular}}
	\end{center}
	
	\vskip -0.2in
	\label{tab:val_set_result}
\end{table*}

\subsection{Deeplabv3+ as a Decoder}
\label{sec:5}
Deeplabv3+~\cite{deeplabv3plus} presented a simplified decoder that is composed of standard $3 \times 3$ convolution and upsampling layers. Here, we added another $3 \times 3$ convolution layer and reduced the number of filters in all $3 \times 3$ convolution from 256 to 96 for computational performance gain. Additional, the output of the encoder is augmented with low level features from earlier layers of the backbone network via long residual connection. These low level features might have large number of feature maps, and in order to resolve this problem, a $1 \times 1$ convolution is utilized to reduce the number of channels of low level feature. Otherwise, with some light backbone networks, there will be no need to apply the $1 \times 1$ convolution on low level features because of the low number of channels (e.g., 24 in case of using MobileNet).


\section{Experimental Results and Validation}
\label{sec:e1}
In our evaluation of the proposed method, the effectiveness of LiteSeg with different backbone networks is empirically tested and the results are compared with the lightweight state-of-the-art architectures on Cityscapes~\cite{dataset} dataset. The performance of the proposed model is measured in terms of mean intersection over union (mIOU), giga floating point operations (GFLOPs), and the number of parameters (Params) in millions.

\subsection{Dataset and Computing Environment}
\label{sec:e2}
The Cityscapes dataset is a large-scale dataset for semantic understanding of urban scenes. It contains 5000 images with fine annotations divided into 2975 images for training, 500 images for validation, and 1525 images for testing. It also contains about 20000 images with coarse annotations that can be used as extra data for fine-tuning the models. 

The experiments were carried out on a computer with Intel Core i7-8700 @ 3.2GHZ, 16GB memory, and NVIDIA GTX1080Ti GPU card. This computer runs Ubuntu 18.04 and PyTorch \cite{pytorch} version 0.4.1 with CUDA 9.0 and cudnn 7.0.5.

\subsection{Training Protocol}
\label{sec:e4}
Stochastic Gradient Descent with Nesterov \cite{momentum} was used with a momentum value of 0.9 and an initial learning rate of $10^{-7}$ for ShuffleNet and MobileNetV2 backbone networks and $10^{-8}$ for Darknet19 backbone network, and a weight decay $4\times 10^{-5}$. We applied multiple learning rate policies where the learning rate changes after every five epochs such that the learning rate of the current epoch is calculated by $\displaystyle initial\_learning\_rate\times(1 - epoch/max\_epochs)^{power}$ with power 0.9.

%
%

\subsection{Encoder Options}
\label{sec:e5}

\textbf{Baseline Model.} First our experiments are conducted with a baseline architecture which employs  ASPP and decoder modules from the Deeplabv3+~\cite{deeplabv3plus}. This baseline was tested with three different backbone networks, MobileNetV2~\cite{mobilenetv2}, ShuffleNet~\cite{shufflenet}, and Darknet19~\cite{yolo} with output stride of 32 during both training and testing phases. As shown in the first row of Table~\ref{tab:val_set_result}, employing Darknet19 as the backbone network for LiteSeg produces an appreciable improvement in the accuracy when compared to MobileNetV2 and ShuffleNet as it is a more efficient classification model~\cite{yolo}. This can be attributed to the fact that the generated features for the decoders make the architecture more efficient as a classifier.

\textbf{\\Employing DASPP Module.} As shown in the second row of Table~\ref{tab:val_set_result}, employing DASPP module along with decreasing output stride from 32 to 16, considerably increases the accuracy of the network by 2.37\% when using Darknet19 as a backbone network. It also shows that employing DASPP module along with keeping output stride at 32 for MobileNetV2 and ShuffleNet increases the accuracy of the network by 0.1\% and 0.9\%, respectively. 
For DASPP module, we employed dilation rates (3,6,9) for the three $3 \times 3$ convolutions in the first layer, and used standard $3 \times 3$ convolution in the second layer of convolutions.

\textbf{Pre-Training on The Coarse Dataset.} Due to the lack of finely annotated data for semantic segmentation models, several works~\cite{fine-tuning1,fine-tuning2} found that object-level and image-level labels can improve the result of semantic segmentation models. LiteSeg is trained on the coarse data for 20 epochs and then the trained model is used for training the fine data. The third row of Table~\ref{tab:val_set_result} shows that using a trained network on coarsely annotated data improves the accuracy of the network by 0.7\%, 1.6\%, and 1.3\% when using Darknet19, MobileNetV2, and ShuffleNet, respectively. 
\begin{table}[h!]
\caption{Effect of employing depthwise separable convolution to reduce the number of floating point operations, instead of standard convolution. The unit of all listed number is Giga Floating Point Operations (GFLOPs). They are measured on image size $1024\times512.$ }
	\begin{center}
	\scalebox{0.73}{
		\begin{tabular}{@{}llllllll@{}}
	\toprule
	Convolution type                & LiteSeg-Darknet  & LiteSeg-MobileNet & LiteSeg-ShuffleNet \\ \midrule
Standard Convolution           & 123.26          & 18.86             & 9.36            \\
Depthwise Separable convolution & 103.09          & 4.9               & 2.75            \\ \bottomrule

		\end{tabular}}
	\end{center}
	
	\label{tab:flops_dw}

	\vskip -0.2in
\end{table}

\textbf{Multi-Scale Input.} Learning network with multi-scale images forces the network to well predict across multiple sizes of input images~\cite{yolo}. Following this strategy, we augmented the dataset with multi-scale input images as our network is a fully convolutional network which makes it accept different dimensions of images. This makes the proposed models efficient for predicting various sizes of input images, as stated in the fourth row of Table~\ref{tab:val_set_result}.

\textbf{Employing Depthwise Separable Convolution.} Not only does the use of depthwise separable convolution instead of using standard convolution in our network reduce the FLOPS as stated in Table~\ref{tab:flops_dw}, but it also improves the accuracy of the network by 0.5\%, 0.7\%, and 0.7\% when using Darknet19, MobileNetV2, and ShuffleNet, respectively when evaluating images of size $1024\times2048$, as stated in the fifth rows of Table~\ref{tab:val_set_result}.

\subsection{Computational Performance Evaluation}
\label{sec:e6}
 The computational efficiency of the proposed models is assessed here. Both the inference time, which reflects the real-time performance, and number of parameters, which reflects the memory footprint, are measured. A set 200 images for the burn-in process and 200 images for evaluation are used in the process. Table~\ref{tab:inference_time} compares the proposed models to current state-of-the-art real-time segmentation networks using the same computing environment. 

\begin{table}[h]
	\caption{Inference time analysis on images with resolution 360x640 and full resolution 1024x2048  using our machine. DSNet result was taken from their paper, they used Nvidia GTX 1080TI on their experiments.}
	\begin{center}
	\scalebox{0.75}{
		\begin{tabular}{@{}llllllll@{}}
	\toprule
	Network             & FPS (360x640) & FPS (1024x2048)& Params(in millions)\\ \midrule  
ErfNet~\cite{erfnet}                          & 105 & 15              & 2.07   \\
DSNet~\cite{dsnet}&100.5& - &0.91\\
LiteSeg-Darknet (ours)            & 98&15  & 20.55  \\  \midrule
ESPNET~\cite{espnet}                   & 144   &25              & 0.364  \\
LiteSeg-MobileNet (ours)            & 161&22  & 4.38   \\
LiteSeg-ShuffleNet (ours)           & 133       &31           & 3.51   \\ \bottomrule
		\end{tabular}}
	\end{center}

	\label{tab:inference_time}

	\vskip -0.2in
\end{table}

These results clearly show the ability of LiteSeg to generate different lightweight models to manipulate the accuracy and computational efficiency by using different backbone network. For example, using $640\times360$ input resolution, LiteSeg with MobileNetV2~\cite{mobilenetv2} as a backbone network achieved a speed of 161 FPS which exceeds the speed of ESPNet~\cite{espnet} by 17 FPS on the same machine, while providing an improved accuracy by 7.51\%.

\subsection{Cityscapes Benchmark Results} 
The models with the best result on the validation set are selected and compared with the results of the proposed model when experimented in the test set. The results are then uploaded to the official benchmark of Cityscapes dataset. As shown in Table~\ref{tab:testset_results}, we compare our result on the test set with other state-of-the-art real-time models for semantic image segmentation. Although the LiteSeg-DarkNet19 has a high GFLOPS compared with ERFNet, it has improved the accuracy of ERFNet and DSNet by 2.75\% and 1.45\%, respectively, just with a sacrifice of 7 FPS for ERFNet and 2.5 FPS for DSNet (Table~\ref{tab:inference_time}). 
\begin{table}[h!]
	\caption{Performance of our propose LiteSeg and similar architectures on Cityscapes test set. For results with '*', GFLOPs is computed on image
resolution 640x360.}
	\begin{center}
	\scalebox{0.9}{
		\begin{tabular}{@{}llllllll@{}}
	\toprule
Model                      & GFLOPS                 & Class mIOU   &Category mIOU    \\ \midrule
SegNet* \cite{segnet}                    & 286.03 & 56.1\%& 79.1\%      \\
ESPNet \cite{espnet}                     & 9.67                   & 60.3\%&82.2\% \\
ENet \cite{enet}                       & 8.52                   & 58.3\% & 80.4\%    \\
ERFNet  \cite{erfnet}                    & 53.48                  & 68.0\% &86.5\%      \\
SkipNet-ShuffleNet \cite{rtseg}          & 4.63                   & 58.3\%&  80.2\%     \\
SkipNet-MobilenetNet \cite{rtseg}     & 13.8                   & 61.5\%  & 82.0\%    \\ 
CCC2 \cite{ccc2}     & 6.29                   & 61.96\%   &  nan  \\ 
DSNet \cite{dsnet}     & nan                   & 69.3\%   & 86.0\%   \\ \midrule
LightSeg-MobileNet (ours)  & 4.9                    & 67.81\% & 86.79\%    \\
LightSeg-ShuffleNet (ours) & 2.75              & 65.17\% &  85.39\%   \\
LightSeg-DarkNet19 (ours)  & 103.09                 & 70.75\%&  88.29\%    \\ \bottomrule
		\end{tabular}}
	\end{center}

	\label{tab:testset_results}
	\vskip -0.2in
\end{table}

\begin{table}[h!]
	\caption{Category results of our LiteSeg models on Cityscapes test set. All number represent the mIOU. }

	\begin{center}
	\scalebox{0.75}{
		\begin{tabular}{@{}llllllll@{}}
	\toprule
Model               & Flat  & Nature & Object & Sky   & Construction & Human & Vehicle \\ \midrule
LightSeg-MobileNet  & 97.90\% & 91.70\%  & 62.75\%  & 94.62\% & 90.44\%        & 79.26\% & 90.88\%   \\
LightSeg-ShuffleNet & 97.88\% & 91.22\%  & 57.43\%  & 93.99\% & 89.69\%        & 77.27\% & 90.28\%   \\
LightSeg-DarkNet19  & 98.44\% & 98.44\%  & 65.94\%  & 94.99\% & 91.67\%        & 81.73\% & 92.95\%   \\ \bottomrule
		\end{tabular}}
	\end{center}
	\label{tab:category_results}
	\vskip -0.2in
\end{table}

\begin{figure*}[t]
\begin{center}
    \includegraphics[width=0.32\textwidth]{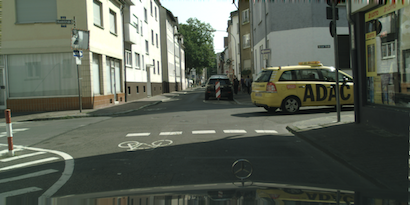}
    \includegraphics[width=0.32\textwidth]{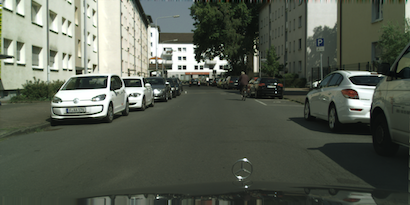}
    \includegraphics[width=0.32\textwidth]{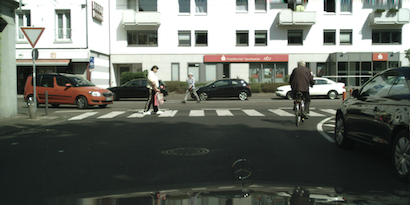}
    \\

    \includegraphics[width=0.32\textwidth]{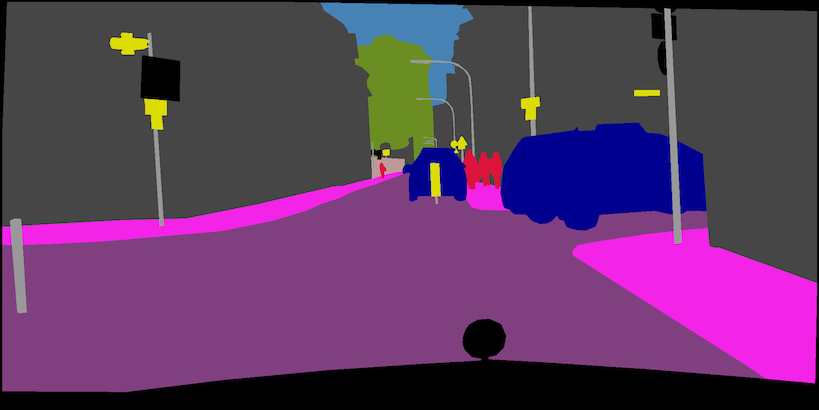}
    \includegraphics[width=0.32\textwidth]{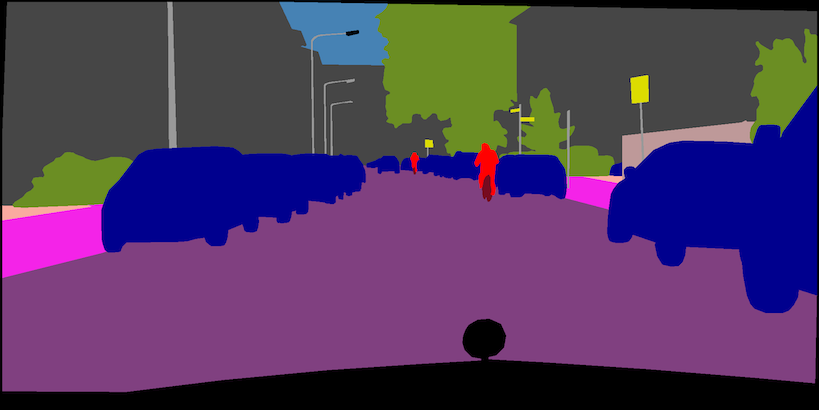}
    \includegraphics[width=0.32\textwidth]{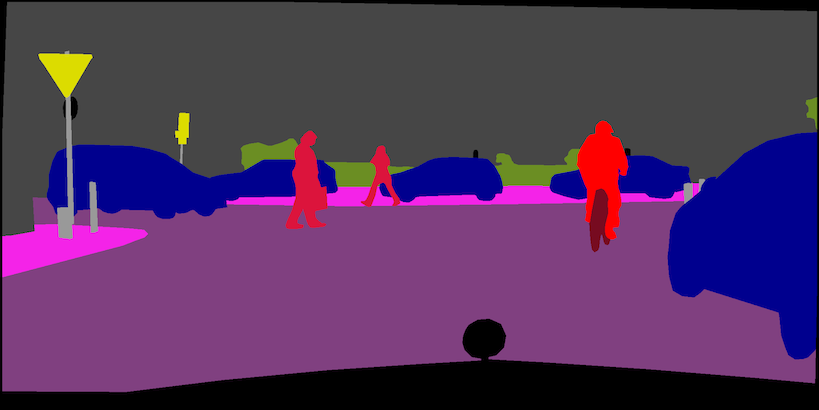}
    \\

    \includegraphics[width=0.32\textwidth]{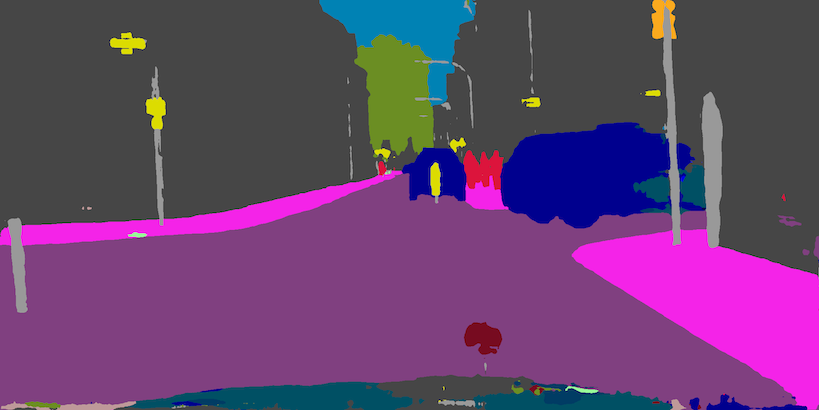}
    \includegraphics[width=0.32\textwidth]{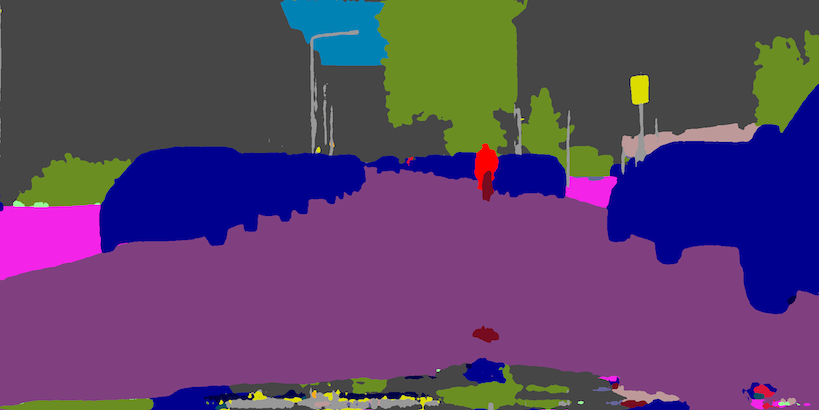}
    \includegraphics[width=0.32\textwidth]{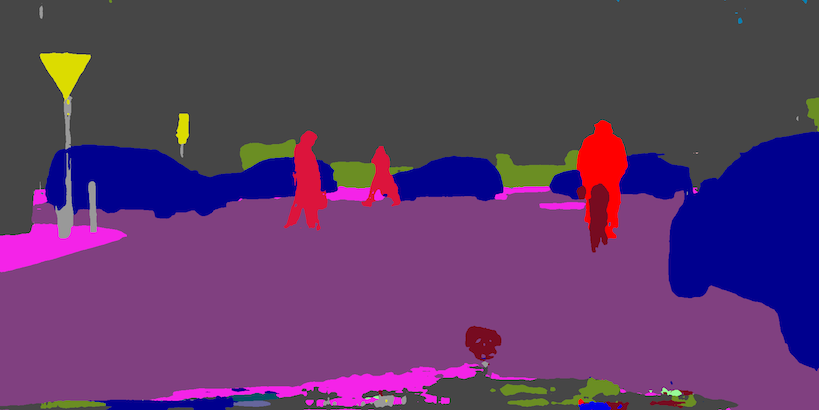}
    \\

    \includegraphics[width=0.32\textwidth]{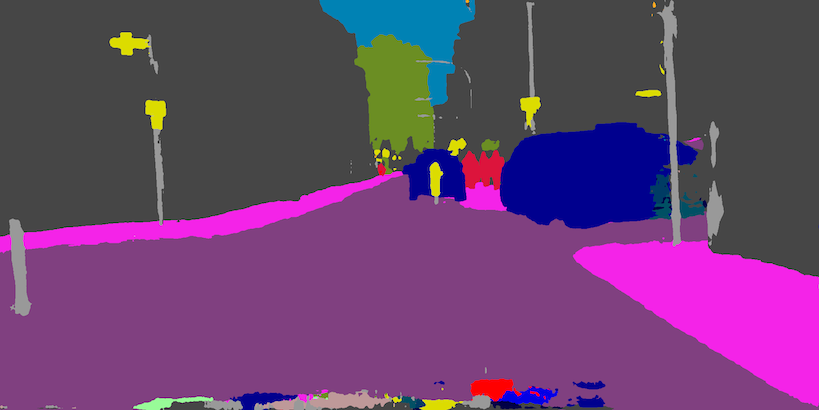}
    \includegraphics[width=0.32\textwidth]{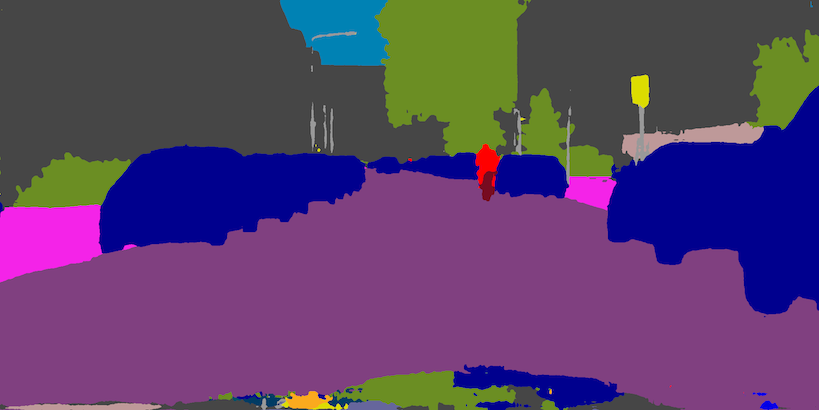}
    \includegraphics[width=0.32\textwidth]{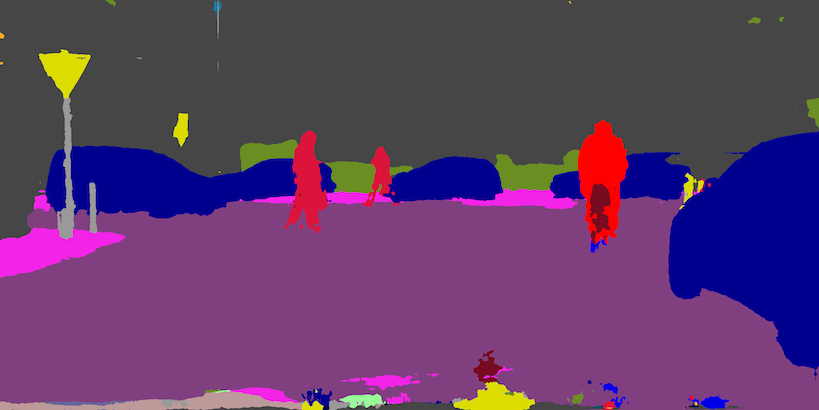}
    \\

    \includegraphics[width=0.32\textwidth]{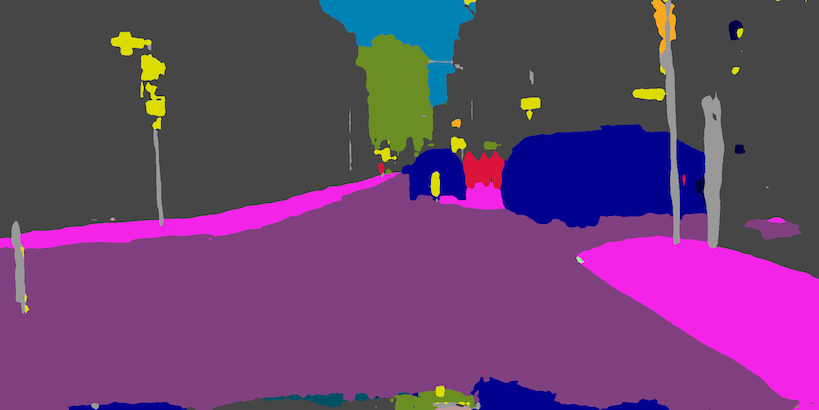}
    \includegraphics[width=0.32\textwidth]{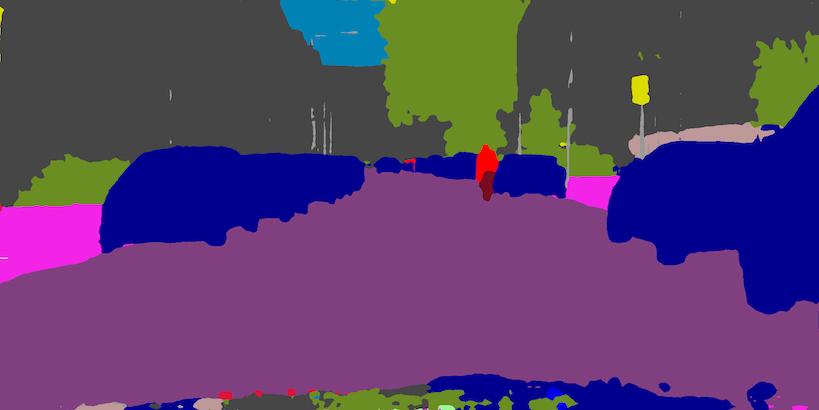}
    \includegraphics[width=0.32\textwidth]{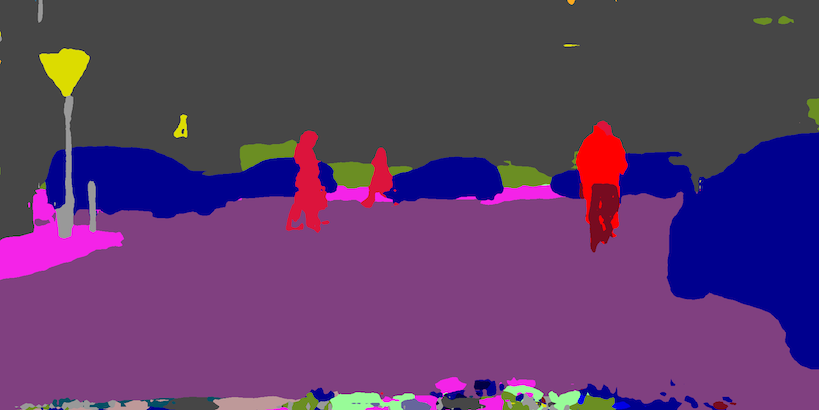}
     \\

    \includegraphics[width=0.32\textwidth]{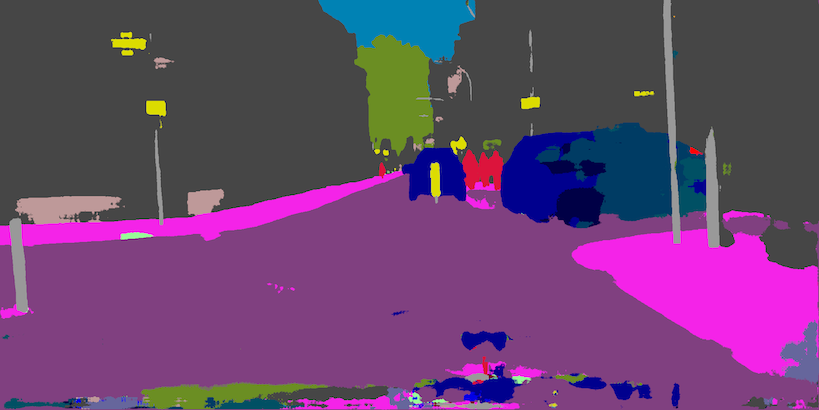}
    \includegraphics[width=0.32\textwidth]{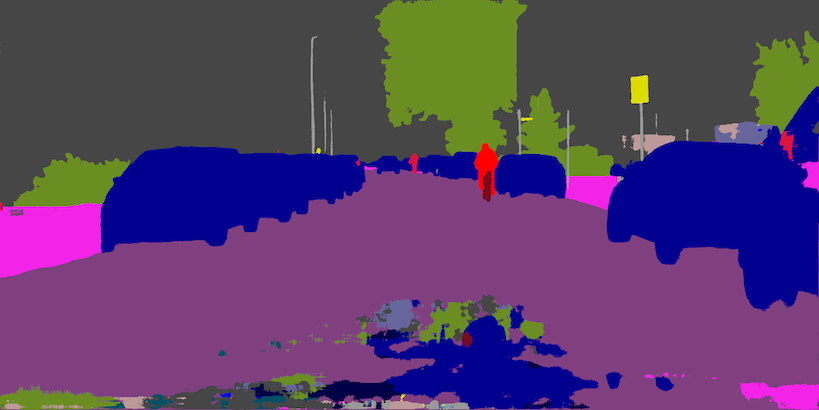}
    \includegraphics[width=0.32\textwidth]{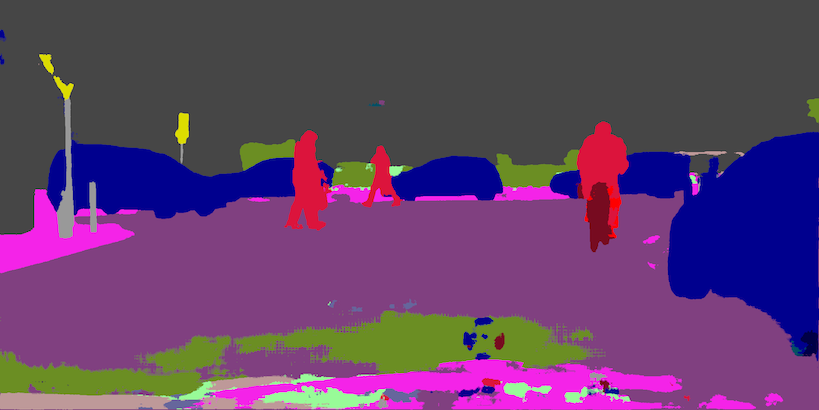}
    \\

    \includegraphics[width=0.32\textwidth]{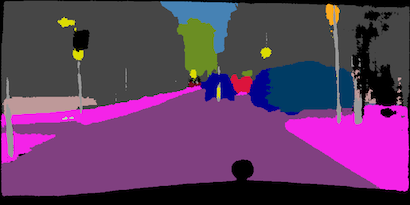}
    \includegraphics[width=0.32\textwidth]{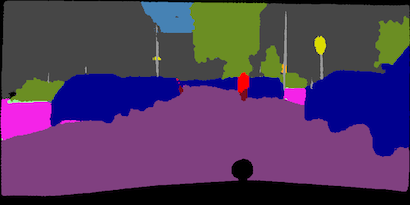}
    \includegraphics[width=0.32\textwidth]{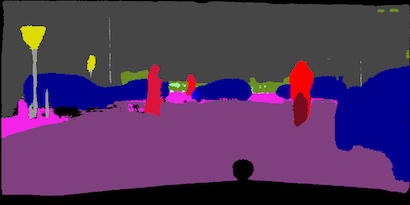}  
    \includegraphics[width=.99\textwidth]{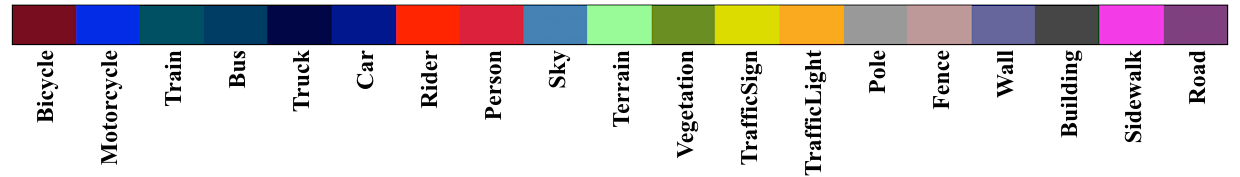} 
\end{center}
\caption{Visualization results of multiple models on Cityscapes validation set~\cite{dataset}. From top to down 1-Input RGB images; 2-Ground truths; 3-LiteSeg-Darknet predictions; 4-LiteSeg-MobileNet predictions; 5-LiteSeg-ShuffleNet predictions; 6-ERFNet predictions; 7-ESPNet predictions; 8- Color map for Cityscapes classes.}
\label{fig:results-cityscapes-viz}
\end{figure*}

In Table~\ref{tab:category_results}, the mIOU of the main categories of Cityscapes test set are listed and one can easily observe that the most common categories in the dataset have the highest mIOU score. The results of LiteSeg are displayed in Figure~\ref{fig:results-cityscapes-viz} for qualitative analysis against ESPNet~\cite{espnet} and ERFNet~\cite{erfnet}.

\section{Conclusion}
In this paper, we proposed LiteSeg, a novel lightweight architecture for semantic image segmentation. The ability of LiteSeg to adapt multiple backbone networks allows for providing multiple trade-offs to fit embedded devices and deep learning workstations. We introduced a new module named DASPP to improve semantic boundaries of captured features from backbone network. The proposed network, LiteSeg, was evaluated with ShuffleNet as a backbone network on the Cityscapes test dataset showing that it is able to achieve 65.17\% mIoU at 31 FPS for full image resolution 1024x2048 on a single Nvidia GTX 1080TI GPU.

\clearpage

\bibliographystyle{ieeetr}
\bibliography{export.bib}


\end{document}